\documentclass{article}
\usepackage{spconf,amsmath,graphicx,hyperref}
\usepackage{amssymb}
\usepackage{booktabs}
\usepackage{multirow}
\usepackage{capt-of}
\usepackage{cite}
\usepackage{amsmath,amssymb,amsfonts}
\usepackage{algorithmic}
\usepackage{graphicx}
\usepackage{textcomp}
\usepackage{xcolor}
\usepackage[table]{xcolor}
\usepackage{xcolor}
\usepackage[normalem]{ulem}
\useunder{\uline}{\ul}{}
\usepackage{float}

\title{Adaptive Visual Token Reduction for Accelerated Image Understanding}
\name{Seyoung Jeong$^{1}$, Jong Pil Yun$^{2,3}$, Sang Jun Lee$^{1*}$,}
\address{
$^{1}$Jeonbuk National University,
Jeonju, Republic of Korea\\
$^{2}$Korea Institute of Industrial Technology
(KITECH), Incheon, Republic of Korea\\
$^{3}$Chung-Ang University, Seoul, Republic of Korea\\
}
\begin{document}
%
\maketitle
\begin{abstract}
Large Vision-Language Models achieve strong VQA performance, but processing high-resolution, information-rich images requires substantial computation, motivating visual token reduction. 
However, existing methods often prune individual tokens or rely on fixed-size cropping, limiting their ability to preserve spatially structured information such as horizontally or vertically elongated text.
To address this limitation, we propose ReFIT, an instruction-guided visual token reduction framework for efficient LVLM inference. 
ReFIT consists of Relevance-Guided Window Reshaping (RWR) and Instruction-Guided Token Refinement (ITR), where RWR captures instruction-relevant regions by adapting to their spatial characteristics, while ITR further removes unnecessary visual tokens. 
Experiments on four VQA benchmarks demonstrate that ReFIT improves answer accuracy while reducing computational cost, and qualitative results demonstrate its effectiveness in localizing relevant regions and removing unnecessary visual information.
\end{abstract}

\vspace{-2mm}
\section{Introduction}
Large Vision-Language Models (LVLMs) have recently demonstrated strong performance across a wide range of multimodal understanding tasks, including Visual Question Answering (VQA)~\cite{alayrac2022flamingo,li2023blip2,bai2023qwen}.
With their enhanced multimodal reasoning capabilities, LVLMs can effectively capture complex relationships between visual content and textual instructions~\cite{liu2023visual,dai2023instructblip}.
Nevertheless, accurately answering questions about information-rich images remains challenging, as relevant evidence is often contained in fine-grained visual content such as small text, tables, and densely arranged elements.
Capturing such details often requires high-resolution inputs, which produce longer visual token sequences and increase inference cost.
Processing all of these tokens is particularly inefficient when the information required to answer a question is contained in only a limited portion of the image. 
Therefore, efficient LVLM inference requires selectively focusing on question-relevant visual information while reducing unnecessary visual tokens.

Existing approaches~\cite{yang2025vflowopt, sun2025lvpruning} improve efficiency by reducing visual tokens through token-level pruning or region-based selection.
Token pruning methods~\cite{sun2025lvpruning,tang2025covipal} estimate token importance and discard less relevant visual tokens, reducing the computational cost of the language model.
However, token-level decisions are limited in preserving the spatial structure of visual content, when semantically related information, such as text, extends across multiple neighboring tokens.
Region-based methods~\cite{jiang2025teva,kwon2026focus} preserve such local structures by identifying question-relevant areas and processing the selected regions in greater detail.
However, existing region selection strategies often rely on fixed-size regions without explicitly considering the spatial characteristics of textual information, limiting the effective learning of question-relevant regions.
Moreover, even correctly identified regions still contain irrelevant information, which degrades answer accuracy.

To address these limitations, we propose ReFIT, a novel approach for efficient LVLM inference.
Given an input image and instruction, our method first identifies instruction-relevant regions using Relevance-guided Window Reshaping (RWR).
RWR captures relevant visual information across diverse spatial configurations without relying on a fixed region structure.
To further refine the selected regions, Instruction-guided Token Refinement (ITR) constructs an instruction-guided relevance map and removes tokens with low relevance to the given instruction.
As a result, ReFIT forwards a compact set of instruction-relevant visual tokens to the language model, improving both answer accuracy and inference efficiency.

\begin{figure*}[!h]
    \centering
    \includegraphics[width=0.95\linewidth]{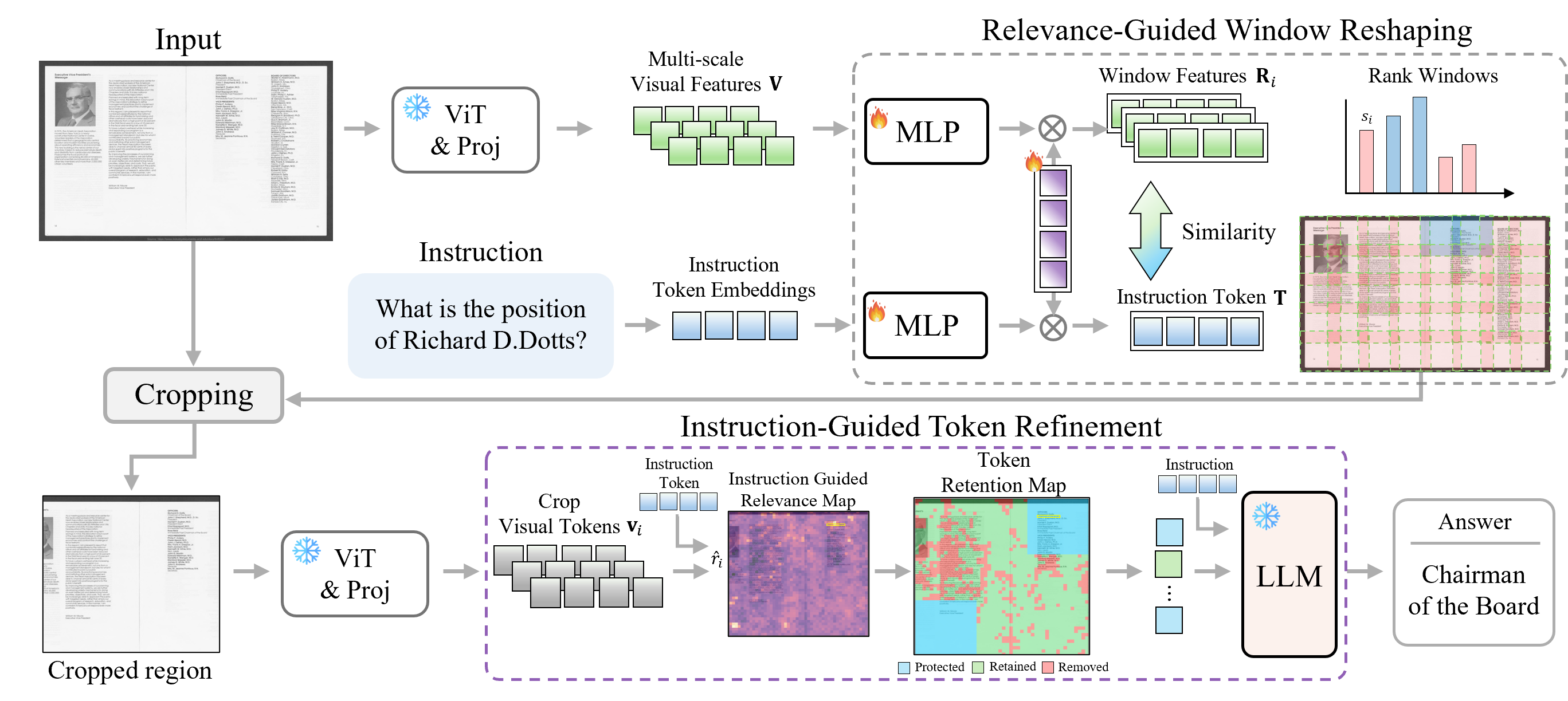}
    \caption{Overall architecture of ReFIT. Given an input image and instruction, RWR adaptively reshapes the window configuration to localize instruction-relevant regions, which are then cropped and re-encoded for fine-grained visual representation. ITR further removes unnecessary visual tokens within the cropped regions using an instruction-guided relevance map, and the remaining tokens are forwarded to the LLM for answer generation.}
    \label{fig:seyoung1}
\end{figure*}

Our contributions are summarized as follows:
\vspace{-1mm}
\begin{itemize}
    \item We propose RWR, which adaptively reshapes the window configuration to capture instruction-relevant regions across diverse spatial layouts.
    \vspace{-1mm}
    \item We propose ITR, which uses an instruction-guided relevance map to remove low-relevance visual tokens while preserving instruction-relevant information.
    \vspace{-1mm}
    \item Extensive experiments on InfoVQA, SPDocVQA, MPDocVQA, and GQA demonstrate that ReFIT consistently improves both answer accuracy and computational efficiency.
\end{itemize}

\section{METHODOLOGY}
We propose ReFIT, a novel approach for efficient LVLM inference, as presented in Fig.~\ref{fig:seyoung1}.
The input image is encoded at multiple resolutions to obtain multi-scale visual features, which are grouped into windows and transformed into window-level representations.
In parallel, the instruction is encoded into instruction-aligned token features.
RWR estimates instruction--window relevance through token-wise similarity and adaptively reshapes the window configuration according to the spatial extent of instruction-relevant regions.
The reconfigured windows are then used to learn instruction-relevant regions, which are subsequently cropped from the original image and re-encoded to obtain refined visual features.
ITR further refines the selected regions using an instruction-guided relevance map, removing visual tokens with low relevance to the instruction.
Finally, the remaining visual tokens are forwarded to the LLM together with the instruction, enabling accurate and efficient answer generation.

\subsection{Feature Extraction}
Given an input image, visual features are first extracted using the pretrained vision encoder and projection layer, yielding
\(\mathbf{V}\in\mathbb{R}^{T\times d}\), where \(T\) denotes the number of visual tokens and \(d\) is the feature dimension.
The visual features are then reshaped into a two-dimensional spatial grid and grouped into local windows, with the feature representation of the \(i\)-th window denoted as \(\mathbf{R}_i\).
In parallel, the instruction is tokenized and embedded to obtain instruction token
\(\mathbf{T}\in\mathbb{R}^{M\times d}\), where \(M\) denotes the number of instruction tokens.
The window features and instruction token are subsequently projected into a shared feature space for relevance estimation.
 
\begin{table*}[th!]
\caption{Quantitative comparison with state-of-the-art methods on VQA benchmarks: InfoVQA, SPDocVQA, MPDocVQA, and GQA}
\label{tab:table1}
\resizebox{\linewidth}{!}{%
\begin{tabular}{c|l|cc|cc|cc|cc}
\hline
 & \multicolumn{1}{c|}{} & \multicolumn{2}{c|}{InfoVQA} & \multicolumn{2}{c|}{SPDocVQA} & \multicolumn{2}{c|}{MPDocVQA} & \multicolumn{2}{c}{GQA} \\ \cline{3-10} 
\multirow{-2}{*}{LLM} & \multicolumn{1}{c|}{\multirow{-2}{*}{Method}} & ANLS↑ & FLOPs (T)↓ & ANLS↑ & FLOPs (T) ↓ & ANLS↑ & FLOPs (T)↓ & Acc↑ & FLOPs (T)↓ \\ \hline
 & Vanilla~\cite{li2024llava} & 0.2552 & 38.98 & 0.6628 & 51.68 & 0.3758 & 50.01 & 0.7598 & 30.39 \\
 & Random   sampling & 0.2387 & 24.62 & 0.4888 & 27.70 & 0.2988 & 26.66 & 0.7484 & 19.26 \\
 & ToMe~\cite{bolya2023token} & 0.1975 & 26.67 & 0.3215 & 39.41 & 0.2135 & 38.32 & 0.7293 & 19.10 \\
 & FastV~\cite{chen2024image} & 0.2306 & 26.22 & 0.6099 & 28.10 & 0.3523 & 29.31 & 0.7478 & 19.23 \\
 & Pdrop~\cite{xing2025pyramiddrop} & 0.2335 & 26.00 & 0.5507 & 26.93 & 0.3637 & 30.82 & 0.7436 & 20.33 \\
 & SparseVLM~\cite{zhang2025sparsevlm} & 0.2428 & 27.45 & 0.5726 & 32.63 & 0.3087 & 31.66 & 0.7449 & 17.85 \\
 & PinPoint~\cite{kwon2026focus} & 0.3024 & 25.48 & 0.6472 & 28.44 & 0.3866 & 28.28 & 0.7608 & 17.96 \\
\multirow{-8}{*}{\begin{tabular}[c]{@{}c@{}}LLaVA\\ -NeXT-7B\end{tabular}} & \cellcolor[HTML]{D9D9D9}Ours & \cellcolor[HTML]{D9D9D9}\textbf{0.3189} & \cellcolor[HTML]{D9D9D9}\textbf{24.58} & \cellcolor[HTML]{D9D9D9}\textbf{0.6638} & \cellcolor[HTML]{D9D9D9}\textbf{28.06} & \cellcolor[HTML]{D9D9D9}\textbf{0.4125} & \cellcolor[HTML]{D9D9D9}\textbf{27.91} & \cellcolor[HTML]{D9D9D9}\textbf{0.7669} & \cellcolor[HTML]{D9D9D9}\textbf{14.94} \\ \hline
\end{tabular}}
\end{table*}

\subsection{Relevance-Guided Window Reshaping}
RWR is designed to overcome the limitation of fixed window configurations in region-based visual selection.
Fixed windows can lead to incomplete learning of instruction-relevant regions when the spatial extent of text or tabular structures exceeds predefined window boundaries.
To address this limitation, RWR adaptively reshapes the window configuration according to instruction relevance, enabling instruction-relevant regions to be learned across diverse spatial layouts.

For each window $\mathbf{R}_i$, composed of visual tokens $\{\mathbf{r}_{i,k}\}_{k=1}^{K_i}$, RWR compares the visual tokens with the instruction tokens $\{\mathbf{t}_j\}_{j=1}^{M}$ to estimate window relevance.
Specifically, the cosine similarity between the $k$-th visual token in the $i$-th window and the $j$-th instruction token is computed as

\begin{equation}
a_{i,k,j}
=
\frac{\mathbf{r}_{i,k}^{\top}\mathbf{t}_j}
{\|\mathbf{r}_{i,k}\|_2 \, \|\mathbf{t}_j\|_2}.
\end{equation}
For each instruction token, the maximum similarity over the visual tokens within the window is selected, and these token-wise scores are aggregated to obtain the relevance score of the $i$-th window:

\begin{equation}
s_i
=
\frac{1}{M}
\sum_{j=1}^{M}
\max_{k} a_{i,k,j}.
\end{equation}

The candidate windows are then ranked according to their relevance scores.
Based on the spatial distribution of high-ranking windows, RWR adaptively adjusts the number of neighboring windows from 2 to 4.
When the instruction-relevant region extends horizontally, adjacent windows are progressively aggregated along the horizontal direction.
For vertically extended regions, neighboring windows are aggregated along the vertical direction.
Through this relevance-guided reshaping, RWR forms region candidates that better match the spatial extent of relevant content and reduce fragmentation caused by fixed window boundaries.

\subsection{Instruction-Guided Token Refinement}
RWR effectively localizes instruction-relevant regions, but the selected regions still contain visual tokens unrelated to the instruction.
To further refine these regions, we propose ITR, which evaluates token-level relevance to the instruction and selectively removes low-relevance visual tokens.
This enables fine-grained token refinement within the selected regions before LLM inference.

Given the refined visual features obtained from the selected regions, ITR computes the relevance of each visual token to the instruction tokens.
Specifically, the cosine similarity between the $i$-th visual token $\mathbf{v}_i$ and the $j$-th instruction token $\mathbf{t}_j$ is computed as

\begin{equation}
s_{i,j}
=
\frac{\mathbf{v}_i^{\top}\mathbf{t}_j}
{\|\mathbf{v}_i\|_2 \|\mathbf{t}_j\|_2}.
\end{equation}
To aggregate the token-level similarities, we normalize the similarities across instruction tokens using a temperature-scaled softmax:

\begin{equation}
\alpha_{i,j}
=
\frac{\exp(s_{i,j}/\tau)}
{\sum_{m=1}^{M}\exp(s_{i,m}/\tau)}.
\end{equation}
The relevance score of each visual token is then computed as

\begin{equation}
r_i
=
\sum_{j=1}^{M}
\alpha_{i,j}s_{i,j}.
\end{equation}
The resulting scores are normalized and mapped back to their spatial locations to form the Instruction-Guided Relevance Map:

\begin{equation}
\hat{r}_i
=
\frac{r_i-r_{\min}}
{r_{\max}-r_{\min}+\epsilon}.
\end{equation}

Based on this map, we construct a Token Retention Map that assigns each visual token to one of three states: \textit{Protected}, \textit{Retained}, or \textit{Removed}.
Tokens belonging to the top-2 regions ranked by RWR are designated as \textit{Protected} and are always preserved.
For the remaining regions, tokens are classified as \textit{Retained} or \textit{Removed} according to their normalized relevance scores.
The final visual token set consists of Protected and Retained tokens, which are forwarded to the LLM for final answer generation.

\section{Experiments}
\subsection{Implementation and Evaluation Details}
We evaluate our method on LLaVA-NeXT-Vicuna-7B~\cite{li2024llava} using four VQA benchmarks: InfoVQA~\cite{mathew2022infographicvqa}, SPDocVQA~\cite{mathew2021docvqa}, MPDocVQA~\cite{tito2023multipage}, and GQA~\cite{hudson2019gqa}.
For InfoVQA, SPDocVQA, and MPDocVQA, performance is evaluated using Average Normalized Levenshtein Similarity (ANLS)~\cite{biten2019stvqa}, while accuracy is reported for GQA.
We additionally report FLOPs to measure the computational efficiency of each method.
The vision encoder, projection layer, and LLM backbone are kept frozen during training, while only the proposed modules are optimized.

\subsection{Quantitative Analysis}
Table~\ref{tab:table1} presents a quantitative comparison with existing token reduction and region-based approaches on InfoVQA, SPDocVQA, MPDocVQA, and GQA.
Our method consistently achieves the best performance across all benchmarks while maintaining substantially lower computational cost than the vanilla LVLM~\cite{li2024llava}.
On MPDocVQA, our method achieves the highest ANLS of 0.4125, improving over the vanilla model by 0.0367 while reducing FLOPs from 50.01 to 27.91 TFLOPs.
On GQA, our method attains the best accuracy of 0.7669 with only 14.94 TFLOPs, corresponding to a 50.8\% reduction in computational cost compared with the vanilla model.
RWR adaptively captures instruction-relevant regions across diverse spatial layouts, while ITR further removes low-relevance visual tokens within the selected regions. 
Therefore, the proposed method reduces visual tokens while enabling more accurate answer generation.

\subsection{Qualitative Analysis}
\begin{figure}[!t]
    \centering
    \includegraphics[width=\linewidth]{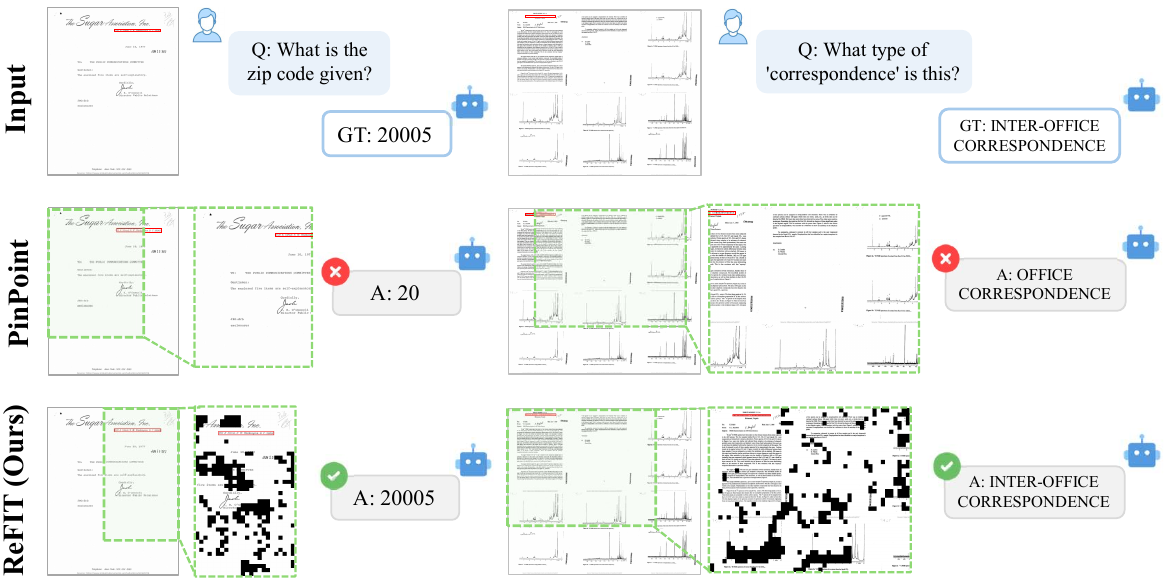}
    \vspace{-7mm}
    \caption{Qualitative comparison of instruction-relevant region cropping and answer generation between existing and proposed methods.  \textcolor{red}{Red boxes} indicate the ground-truth regions, while \textcolor{green!75!black}{green boxes} indicate the regions cropped by each model. The proposed method better localizes horizontally extended text and further removes unnecessary visual tokens within the cropped regions.}
    \label{fig:seyoung2}
\end{figure}

Fig.~\ref{fig:seyoung2} presents a qualitative comparison between the baseline and our method.
RWR first crops instruction-relevant regions, and ITR further removes irrelevant visual tokens within the selected regions.
Compared with PinPoint~\cite{kwon2026focus}, our method better captures the spatial characteristics of horizontally elongated text and more accurately crops instruction-relevant regions, while further reducing unnecessary visual tokens.
In contrast, PinPoint is trained with fixed-size windows, often resulting in truncated horizontally extended text and incorrect answers. Overall, ReFIT accurately localizes instruction-relevant regions while effectively reducing unnecessary visual tokens, thereby enabling more accurate answer generation.

\vspace{-4mm}

\begin{table}[t]
\centering
\caption{Ablation study on the effectiveness of RWR and ITR.}
\resizebox{0.9\columnwidth}{!}{%
\begin{tabular}{c|c|c|c|c|c}
\hline
\multirow{2}{*}{RWR} &
\multirow{2}{*}{ITR} &
\multicolumn{1}{c|}{InfoVQA} &
\multicolumn{1}{c|}{SPDocVQA} &
\multicolumn{1}{c|}{MPDocVQA} &
\multicolumn{1}{c}{GQA} \\ \cline{3-6}

& &
ANLS $\uparrow$ &
ANLS $\uparrow$ &
ANLS $\uparrow$ &
Acc $\uparrow$ \\ \hline

& &
0.3024 &
0.6472 &
0.3866 &
0.7608 \\

\checkmark & &
0.3058 & 0.6529 & 0.3948 & 0.7642 \\

  & \checkmark 
& 0.3077 & 0.6552 & 0.3907 & 0.7612 \\

\checkmark & \checkmark &
0.3189 & 0.6638
& 0.4125 & 0.7669 \\ \hline

\end{tabular}%
}
\label{tab:ab1}
\end{table}

\begin{table}[t]
\centering
\caption{Ablation study on region selection strategies.}
\setlength{\tabcolsep}{2.5pt}
\resizebox{\columnwidth}{!}{%
\begin{tabular}{l|cc|cc|cc|c}
\hline
\multirow{2}{*}{Region selection} &
\multicolumn{2}{c|}{InfoVQA} &
\multicolumn{2}{c|}{SPDocVQA} &
\multicolumn{2}{c|}{MPDocVQA} &
\multicolumn{1}{c}{GQA} \\ \cline{2-8}

& ANLS $\uparrow$ & Acc $\uparrow$
& ANLS $\uparrow$ & Acc $\uparrow$
& ANLS $\uparrow$ & Acc $\uparrow$
& Acc $\uparrow$ \\ \hline

Single best window
& 0.3077 & 0.2535
& 0.6514 & 0.5431
& 0.3994 & 0.3019
& 0.7532 \\

Random multi-window
& 0.3179 & 0.2667
& 0.6643 & 0.5524
& 0.4077 & 0.3071
& 0.7526 \\

Fixed Top-$K$ IoU
& 0.3186 & 0.2596
& 0.6552 & 0.5487
& 0.4060 & 0.3044
& 0.7569 \\

RWR (Ours)
& 0.3189 & 0.2645
& 0.6638 & 0.5562
& 0.4125 & 0.3112
& 0.7669 \\ \hline

\end{tabular}%
}
\label{tab:ab2}
\end{table}

\begin{table}[t]
\centering
\caption{Ablation study on token selection methods.}
\setlength{\tabcolsep}{2.5pt}
\resizebox{\columnwidth}{!}{%
\begin{tabular}{l|cc|cc|cc|c}
\hline
\multirow{2}{*}{Token selection} &
\multicolumn{2}{c|}{InfoVQA} &
\multicolumn{2}{c|}{SPDocVQA} &
\multicolumn{2}{c|}{MPDocVQA} &
\multicolumn{1}{c}{GQA} \\ \cline{2-8}

& ANLS $\uparrow$ &  Acc $\uparrow$
& ANLS $\uparrow$ &  Acc $\uparrow$
& ANLS $\uparrow$ &  Acc $\uparrow$
& Acc $\uparrow$ \\ \hline

Random
& 0.3180 & 0.2638
& 0.6634 & 0.5554
& 0.4119 & 0.3104
& 0.7567 \\

Spatially Connected
& 0.3185 & 0.2642
& 0.6638 & 0.5562
& 0.4120 & 0.3106
& 0.7568 \\

ITR (Ours)
& 0.3189 & 0.2645
& 0.6638 & 0.5562
& 0.4125 & 0.3112
& 0.7669 \\ \hline

\end{tabular}%
}
\label{tab:ab3}
\end{table}

\subsection{Ablation study}
We report ANLS and Region Accuracy on InfoVQA~\cite{mathew2022infographicvqa}, SPDocVQA~\cite{mathew2021docvqa}, and MPDocVQA~\cite{tito2023multipage}, and accuracy on GQA~\cite{hudson2019gqa}.

\noindent\textbf{Ablation Study of RWR and ITR.}
We evaluate the contribution of RWR and ITR through a component-wise ablation study.
As shown in Table~\ref{tab:ab1}, both RWR and ITR individually improve performance, while combining them yields the best results across all benchmarks.
RWR better captures instruction-relevant regions by considering their spatial structures, while ITR further removes unnecessary visual tokens, leading to more accurate answer generation.

\noindent\textbf{Effectiveness of RWR.}
To verify the effectiveness of RWR, we compare various region selection strategies.
As shown in Table~\ref{tab:ab2}, RWR achieves the best overall performance across the four benchmarks.
Compared with fixed or random region selection, RWR adaptively learns window configurations based on text spatial characteristics, enabling more accurate localization of instruction-relevant regions.

\noindent\textbf{Effectiveness of ITR.}
To evaluate the effectiveness of ITR, we compare various token selection methods within the cropped regions.
As shown in Table~\ref{tab:ab3}, ITR achieves the best overall performance across the four benchmarks.
Compared with random or spatially connected token removal, ITR more effectively removes unnecessary visual information using the Instruction-Guided Relevance Map.

\section{Conclusion}
In this work, we propose a framework to accelerate LVLM inference for information-rich image understanding. 
Our method first identifies instruction-relevant regions using Relevance-Guided Window Reshaping (RWR), which adapts the window configuration to diverse spatial layouts. Instruction-Guided Token Refinement (ITR) then removes irrelevant visual tokens within the selected regions, yielding a compact set of instruction-relevant visual evidence. Experiments on InfoVQA, SPDocVQA, MPDocVQA, and GQA show that our method improves both answer accuracy and computational efficiency, even surpassing the vanilla LVLM with substantially lower computation. Overall, the results validate the effectiveness of RWR and ITR for efficient LVLM inference.

\vfill\pagebreak




\bibliographystyle{IEEEbib}
\bibliography{strings,refs}

\end{document}